\documentclass[11pt]{article}

\usepackage[preprint]{acl}

\usepackage{times}
\usepackage{latexsym}
\usepackage[T1]{fontenc}
\usepackage[utf8]{inputenc}
\usepackage{microtype}
\usepackage{inconsolata}
\usepackage{graphicx}
\usepackage{amsmath}

\hypersetup{
  pdftitle={Squeezing More from Limited Data with Recursive Transformers},
  pdfauthor={Serdar G\"ulbahar, Lukas Edman, Alexander Fraser}
}

\title{Squeezing More from Limited Data with Recursive Transformers}

\author{%
  Serdar Gülbahar \qquad Lukas Edman \qquad Alexander Fraser \\
  Technical University of Munich
}

\begin{document}
\maketitle

\begin{abstract}\sloppy
Pre-training under limited data requires a different view of scaling than web-scale language modeling. With a fixed data budget but relatively abundant compute, increasing parameter count helps only up to an optimal scale; beyond that point, models overfit and generalization worsens. We study this behavior across 10M–100M word pre-training budgets, two corpora, and multiple downstream evaluations, and find that optimal size depends strongly on both the data budget and the downstream target. We argue that standard Transformers scale down poorly to this setting, because embeddings consume a large fraction of the parameter budget and per-token computation is tied to representational capacity.
To address this coupling, we study recursive Transformers, reusing a shared block across depth to scale compute, together with factorized embeddings to reduce vocabulary-map parameters. We train three recursive models and find that they outperform standard Transformers at 10M and 100M words, while remaining competitive with BabyLM Challenge 2025 winners.
\end{abstract}

\section{Introduction}
Most work on language model pre-training has been developed in the web-scale regime, where performance is often improved by scaling model size, compute, and data together \citep{kaplan2020scalinglawsneurallanguage,hoffmann2022trainingcomputeoptimal}. Here, we consider a different setting: pre-training under a limited data budget but relatively abundant compute, where web-scale scaling intuitions do not carry over. In limited-data training, more parameters cannot be paired with more data, so increasing model size changes the bias--variance tradeoff. This makes model size itself a regularization choice.

This setting is scientifically interesting because it is closer to developmentally motivated data budgets \citep{hu-etal-2024-findings}, and practically important because continued scaling of foundation models may be constrained by the supply of high-quality human-generated text \citep{10.5555/3692070.3694094}.

Human language learners are exposed to far less linguistic input than today's language models, and the BabyLM Challenge is designed around this idea by evaluating pre-training under developmentally motivated budgets such as 10M or 100M words \citep{gilkerson2017mapping,hu-etal-2024-findings}. Naturally, it is a controlled testbed for us to study how models should be designed when data is scarce.

We argue that standard Transformers scale down poorly to this setting. Scaling down standard Transformers reduces parameter count, but it also changes the balance among vocabulary maps, block capacity, and computational depth. At small scales, a fixed vocabulary size means embeddings and the LM head can dominate the parameter budget, and the computation applied to each token can become too limited. This couples per-token computation to representational capacity, even though in this setting we would like to control them separately. We therefore explore architectural ways to break this coupling: factorized embeddings reduce the parameters spent on vocabulary maps, while recursive weight sharing lets us increase computational depth without a proportional increase in parameter count. We refer to the resulting model family as RecursiveGPT.

We study the following research questions:
\begin{itemize}
    \item What sort of bottlenecks make standard language model architectures scale down poorly to limited-data pre-training budgets?
    \item How does optimal model size depend on the data budget and downstream evaluation target?
    \item Can recursive Transformers provide a better path for scaling compute under limited data?
\end{itemize}
To study these questions, we run experiments spanning data budgets from 10M to 100M words and model sizes from 13M to 1.2B parameters, across two corpora and multiple benchmarks. Our results show that data-constrained pre-training behaves differently from the web-scale regime: the optimal model size depends strongly on both the amount of data and the downstream evaluation, and standard architectures scale down poorly. 

Our central claim is that limited-data pre-training benefits from architectures that decouple representational capacity from per-token computation; our goal is not to improve compute efficiency, but to provide another compute scaling axis that lets us squeeze out more performance out of limited data. RecursiveGPT provides a simple instance of this principle. Empirically, RecursiveGPT uses this additional computation to outperform standard Transformers at both 10M and 100M word budgets, while remaining competitive with BabyLM Challenge 2025 winners. 

We view RecursiveGPT as a useful architectural pathway for scaling up compute under fixed data, while remaining orthogonal to more specialized data-efficient training methods, such as those used in BabyLM. We openly release the training pipeline.\footnote{\href{https://github.com/serdardoesml/recursive-lm}{\nolinkurl{github.com/serdardoesml/recursive-lm}}}

\section{Background}
We focus on the regime where data, rather than computation, is the binding constraint. With a fixed corpus but relatively abundant compute, parameter count becomes a form of regularization: larger models can fit the training corpus more easily, so the natural scaling direction is often to reduce the number of trainable parameters. This creates a tension with compute. The same knobs that usually buy more computation, such as larger width, greater depth, or more passes over the data, also tend to increase the risk of overfitting. At the BabyLM scale, this compute-abundant view is not merely hypothetical. In the 10M-word setting, competitive models can be trained in minutes on a single GPU (see Appendix~\ref{app:compute-cost}), making it natural to ask, in the spirit of ``The Bitter Lesson'' \citep{sutton2019bitterlesson}, how to allocate more compute.

Another layer to this tension is scaling down standard language model architectures. If vocabulary size is held fixed, reducing the hidden dimension shrinks the Transformer blocks roughly quadratically, while the embedding matrix shrinks only linearly. As a result, a small language model can spend most of its parameters on token embeddings rather than on the computation performed at each layer.

Prior work has explored a training-dynamics approach to this problem, keeping models overparameterized while controlling overfitting through aggressively tuned training recipes, including weight decay far above standard pre-training practice and, in later stages, ensembles of independently trained models \citep{kim2026pretraining}. We instead explore an architectural approach to solving this problem, designing the model itself to be better-matched to the data-constrained regime.

\subsection{BabyLM Challenge}
The BabyLM Challenge \citep{hu-etal-2024-findings} studies language model pre-training under developmentally plausible data budgets. Its standard tracks restrict training to at most 10M or 100M words for 10 epochs, but do not restrict FLOPs; making it a useful benchmark for the limited-data setting we consider.

The 2025 challenge ranked systems separately by track and by NLP-task versus human-likeness metrics. The NLP track is most relevant for our evaluations; its strict-small winner, AMLM hard decay, is a masked language model that adapts the masking distribution during training \citep{edman-fraser-2025-mask}. Other strong systems combined causal and masked objectives or used diffusion-style masked language modeling \citep{charpentier-samuel-2024-bert,kosmopoulou-etal-2025-masked,charpentier-etal-2025-findings}.

Two evaluations used by the challenge that we also use later are BLiMP \citep{warstadt-etal-2020-blimp-benchmark} and COMPS \citep{misra-etal-2023-comps}. For comparisons to BabyLM 2025 models later, we also report BLiMP Supplement, EWoK \citep{ivanova-etal-2025-elements}, and Entity Tracking \citep{kim-schuster-2023-entity}, which we treat as supplementary because they appear to be noisier and are less central to our research questions. 

\subsection{Recursive Transformers}
A simple way to untie per-token computation from representational capacity is to share weights across depth. Recursive Transformers do this by replacing a stack of independently parameterized blocks with repeated applications of a shared transition function. Increasing recurrent depth therefore adds computation between the embedding layer and the output head without adding another full set of block parameters. This idea connects several strands of prior work.

The Universal Transformer applies a self-attentive transition recurrently across depth, optionally with adaptive halting, and was motivated by combining the parallelism of Transformers with a recurrent inductive bias \citep{dehghani2019universal}. ALBERT later showed that factorized embeddings and cross-layer parameter sharing can greatly reduce the parameter count of BERT-style encoders \citep{DBLP:conf/iclr/LanCGGSS20}. Deep Equilibrium Models push weight-tied depth further by solving for the fixed point of an effectively infinite-depth network, including self-attention variants \citep{10.5555/3454287.3454350}. More recent recursive-Transformer work has used layer tying and lightweight depth-specific adaptation to compress pretrained language models \citep{bae2025relaxed}, while depth-recurrent language models, such as AbbIE and Huginn, study recurrent unrolling as a way to scale latent test-time computation \citep{aleksandrov2025abbie,geiping2026scaling}.

Recent small recursive reasoning models illustrate the same low-data principle outside language modeling. HRM and TRM solve supervised puzzle tasks by applying compact recurrent modules repeatedly to achieve strong generalization from small training sets \citep{wang2025hierarchicalreasoningmodel,jolicoeurmartineau2025morerecursivereasoningtiny}.

\subsection{Factorized Embeddings}
\label{sec:factorized-embeddings}
Factorized embedding parameterization unties the embedding size $E$ from the Transformer hidden size $H$. In the standard parameterization, the input embedding matrix has size $V \times H$, where $V$ is the vocabulary size. ALBERT introduced factorized embeddings for parameter efficiency, replacing this matrix with two smaller matrices of size $V \times E$ and $E \times H$ \citep{DBLP:conf/iclr/LanCGGSS20}. Equivalently, tokens are first mapped into an $E$-dimensional embedding space and are then projected into the $H$-dimensional hidden space used by the Transformer. Concretely, for a one-hot token vector $x_t$, we compute
\[
\mathrm{Embed}_{\mathrm{FE}}(x_t) = x_t W_{\mathrm{emb}} W_{\mathrm{proj}}.
\]
Here $\mathrm{Embed}_{\mathrm{FE}}$ denotes the factorized embedding map, $W_{\mathrm{emb}} \in \mathbf{R}^{V \times E}$ is the token embedding matrix, and $W_{\mathrm{proj}} \in \mathbf{R}^{E \times H}$ projects into the Transformer hidden space. We use the same factorization for the untied language-model head, mapping a hidden state $h_t$ to vocabulary logits by
\[
\mathrm{Head}_{\mathrm{FE}}(h_t) = h_t W_{\mathrm{proj}} W_{\mathrm{unembed}},
\]
where, for the output head, $W_{\mathrm{proj}} \in \mathbf{R}^{H \times E}$ and $W_{\mathrm{unembed}} \in \mathbf{R}^{E \times V}$. When $E \ll H$, each factorized vocabulary map uses $O(VE + EH)$ parameters instead of $O(VH)$ while leaving the hidden size unchanged. This lets us prevent vocabulary maps from dominating the parameter count at very small scales by reducing $E$ independently of $H$. When we choose $E=H$, we set $W_{\mathrm{proj}}=I_H$, recovering the standard embedding or output-head parameterization.

\begin{figure}[t]
\centering
\includegraphics[width=1\linewidth]{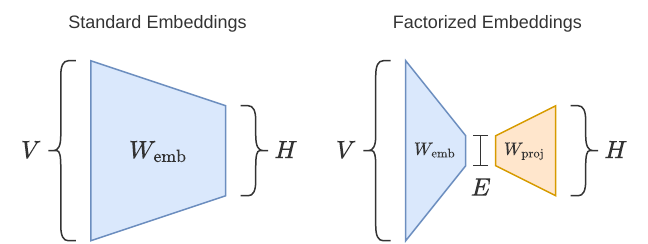}
\caption{Standard embeddings use a single $V \times H$ matrix, while factorized embeddings introduce an embedding-size bottleneck by mapping tokens through an $E$-dimensional space before projecting to the $H$-dimensional hidden space. This reduces embedding parameter count when $E \ll H$ and can act as a useful regularizer.}
\label{fig:factorized-embeddings}
\end{figure}

\begin{figure*}[!t]
\centering
\includegraphics[width=\textwidth]{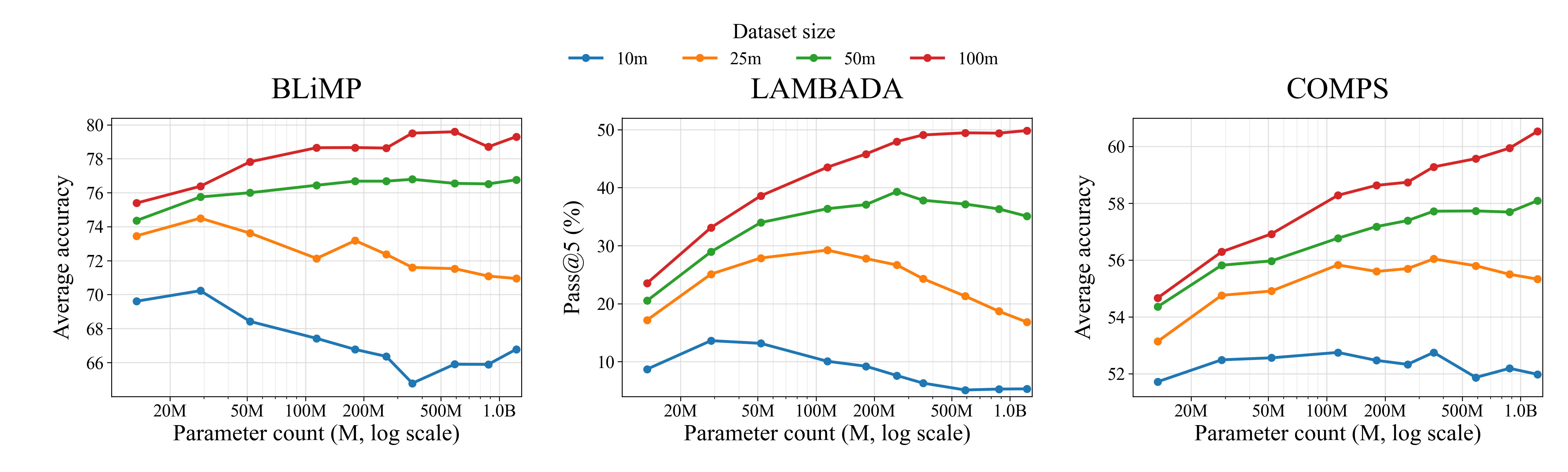}
\caption{Optimal scale with standard Transformers on the BabyLM baseline corpus across three representative evaluations. The preferred parameter count depends on the downstream target: BLiMP saturates relatively early, LAMBADA becomes more scale-favoring as dataset size grows, and COMPS continues to benefit from larger models at moderate and high data budgets.}
\label{fig:optimal-scale}
\end{figure*}

\section{Optimal Scale for Standard Transformers Under Data Scarcity}
\label{sec:optimal-scale}

\subsection{Scaling Experiment Setup}
We first study optimal scale with standard, non-recursive Transformers. All models use 12 layers, and we sweep hidden sizes from 256 to 3072. To keep embeddings from dominating the parameter budget at the smallest scales, we use factorized embeddings for the three smallest models, setting $E$ to 96, 192, and 352 for $H = 256$, 384, and 512, respectively. The $H{=}384$, $E{=}192$ pairing is supported by the factorized-embedding sweep in Appendix~\ref{app:factorized-embedding-sweep}. For all larger models, we set $E = H$. We train on 10M, 25M, 50M, and 100M-word budgets for 10 epochs using two corpora: the BabyLM baseline corpus used by GPT-BERT \citep{charpentier-samuel-2024-bert}, and Nemotron-ClimbMix \citep{diao2026nemotronclimb}. Every 10M words corresponds to roughly 13M BPE tokens. Within each corpus, smaller budgets are constructed as nested, shuffled subsets of larger ones, so scaling dataset size does not intentionally change the data mixture. We defer the exact architecture specification to Appendix~\ref{app:architecture-details}.

We report three representative evaluations in Figure~\ref{fig:optimal-scale}: BLiMP \citep{warstadt-etal-2020-blimp-benchmark}, LAMBADA pass@5 \citep{paperno-etal-2016-lambada}, and COMPS \citep{misra-etal-2023-comps}. BLiMP and COMPS are evaluated using the BabyLM evaluation pipeline \citep{babylm-evaluation-pipeline-2025}. BLiMP is a set of 67 English minimal-pair datasets targeting grammatical phenomena in syntax, morphology, and semantics. COMPS is a minimal-pair benchmark for conceptual property knowledge and property inheritance, testing whether models assign plausible properties to concepts and generalize them across category relations. LAMBADA is a benchmark for broad discourse understanding in which models must predict a missing final word from a passage-level context. We evaluate it using pass@5 under gold-prefix scoring: at each continuation position, conditioned on the true preceding tokens, the reference token must appear among the model's top-5 predictions; multi-token continuations must satisfy this at every position.

\subsection{Scaling Behavior by Evaluation}

Performance is clearly non-monotonic in parameter count, but the pattern differs substantially across evaluations, as illustrated by Figure~\ref{fig:optimal-scale}. BLiMP improves with scale and then mostly saturates, with the clearest signs of overfitting appearing at the smallest data budgets. In contrast, COMPS continues to benefit from larger parameter counts across much of the range we test, especially at 50M and 100M words, including in regions where BLiMP and low-budget LAMBADA have already flattened or begun to decline.

\begin{figure}[t]
\centering
\includegraphics[width=\linewidth]{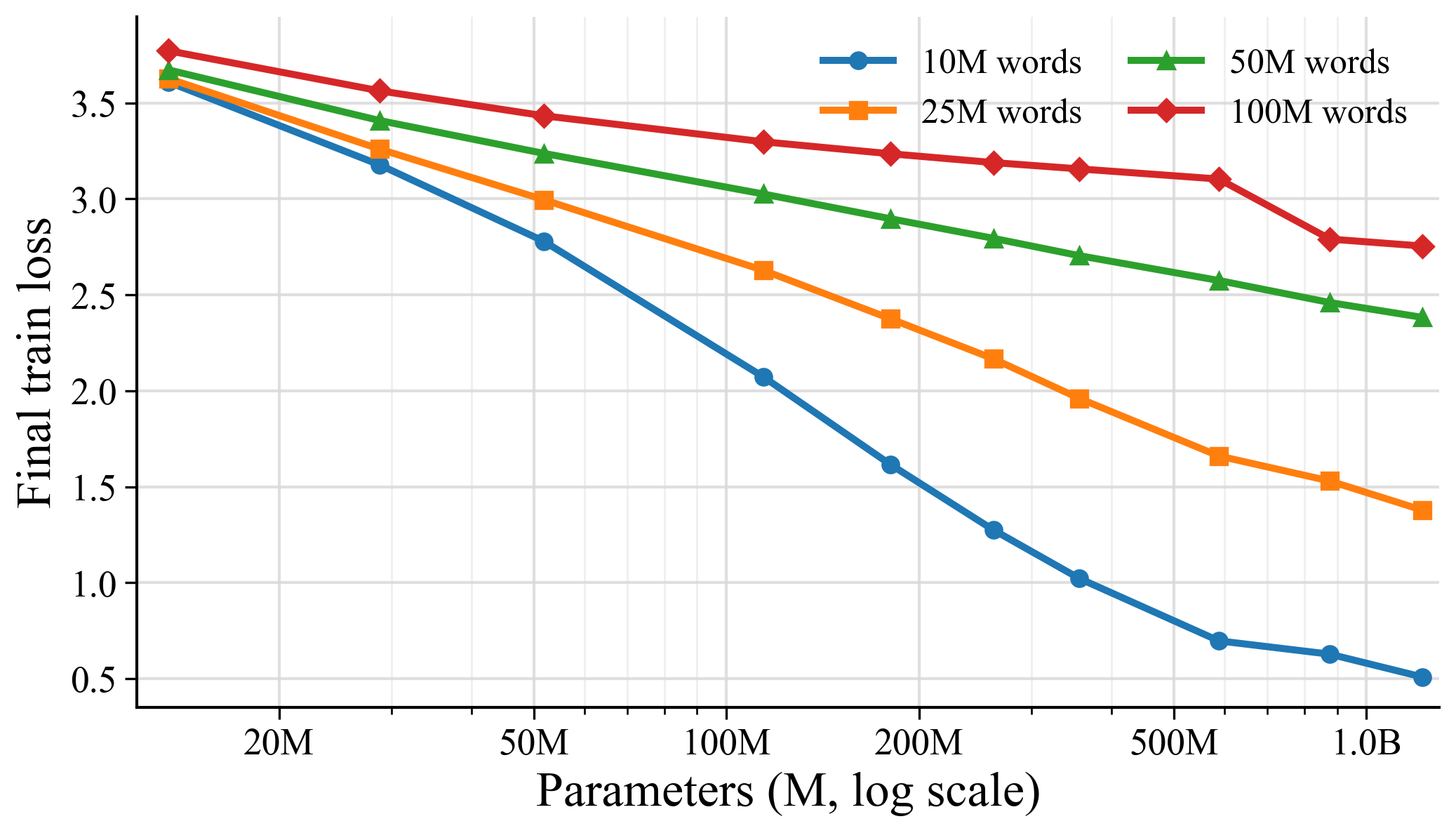}
\caption{Final training loss for the standard-model optimal-scale sweep on the BabyLM baseline corpus. The x-axis uses the same parameter counts as Figure~\ref{fig:optimal-scale}. All data subsets use the same tokenizer and vocabulary, so losses are directly comparable. Training loss keeps falling even when downstream performance saturates or declines.}
\label{fig:optimal-scale-train-loss}
\end{figure}

One plausible explanation is that COMPS tests conceptual property knowledge and property inheritance: the model must encode associations between concepts and properties and apply them across category relations. This makes the task more dependent on memorized semantic knowledge from the pre-training corpus. Larger models may therefore continue to improve on COMPS because additional parameters provide useful associative capacity even after broader generalization gains have tapered off. This again shows that the optimal scale depends strongly on the downstream target.

LAMBADA asks the model to predict a held-out word from the preceding context. Its examples are constructed so that the target word is predictable from the semantic context of the full passage, but not from short-range cues alone. This format remains close to the language-modeling objective, which may explain why its scaling curves are comparatively smooth.

The corresponding final training losses decrease steadily with parameter count (Figure~\ref{fig:optimal-scale-train-loss}), so the downstream declines at small data budgets occur despite better training fit and are consistent with overfitting rather than optimization failure.

The same qualitative picture appears on ClimbMix: preferred parameter ranges remain broadly stable despite shifts in absolute scores, suggesting that scale is determined more by data budget and evaluation target than by corpus choice (Appendix Figure~\ref{fig:optimal-scale-climbmix}).

These results reveal a central obstacle for compute scaling under fixed data. Once model size exceeds the data-dependent optimum, spending additional compute by adding parameters no longer reliably improves generalization and can instead amplify overfitting. Progress in this regime may therefore benefit from an alternative scaling axis: increasing per-token computation without proportionally increasing the number of trainable parameters.

\section{RecursiveGPT Architecture}
\label{sec:recursivegpt-architecture}
We use recursive weight sharing as an architectural response to this compute-scaling problem. The architecture is a recursive causal decoder inspired by ALBERT \citep{DBLP:conf/iclr/LanCGGSS20}, in which a shared Transformer block is applied recurrently across depth. Due to optimization difficulties with recursive weight sharing, and also to explicitly control depth, we keep the design deliberately simple and use a fixed number of recurrent steps. We return to these tradeoffs in Section~\ref{sec:discussion}. We refer to this model family as RecursiveGPT. Our approach is illustrated in Figure~\ref{fig:recursivegpt}. Following the Universal Transformer view of depth recurrence \citep{dehghani2019universal}, we describe the model as repeated application of a single causal decoder transition, initialized by the factorized embedding map and read out with the factorized head from Section~\ref{sec:factorized-embeddings}:

\begin{figure}[t]
\centering
\hspace*{0.2\linewidth}\includegraphics[width=0.8\linewidth]{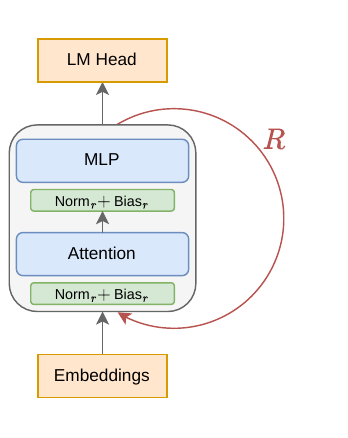}
\caption{RecursiveGPT architecture. Tokens are processed by applying a shared Transformer block $R$ times. Each recurrent step $r$ uses separate normalization and bias parameters for depth conditioning.}
\label{fig:recursivegpt}
\end{figure}

\[
\begin{aligned}
h^{(0)} &= \mathrm{Embed}_{\mathrm{FE}}(x_t), \\
h^{(r)} &= F_{\theta}(h^{(r-1)}; \phi_r), \quad r=1,\ldots,R, \\
y_t &= \mathrm{Head}_{\mathrm{FE}}(h^{(R)}).
\end{aligned}
\]
Here $F_{\theta}$ is the shared Transformer block, $\theta$ contains the attention and MLP weights reused at every recurrent step, $\phi_r$ denotes the lightweight step-specific normalization parameters, and $y_t$ denotes the next-token logits produced by the factorized language-model head. In contrast to a standard $R$-layer Transformer, increasing $R$ therefore adds computation without introducing another full set of block parameters.

Each step still has its own parameterized RMSNorm modules for the attention sublayer, MLP sublayer, and QK normalization, with additional learned biases on the first two. These per-step normalization parameters act as lightweight depth conditioning, and the learned biases in particular function as a lightweight depth embedding, analogous to the recurrent-step timing signal used by \citet{dehghani2019universal}, while most of the block parameters are shared.

Unless otherwise stated, RecursiveGPT uses the same architectural and training details as the standard models described in Appendix~\ref{app:architecture-details}; the key architectural change is that the recurrent block shares weights across depth. One exception is the feed-forward sublayer: we use an unusually high expansion factor of 16, because MLP neuron count scales linearly with hidden size whereas total parameter count scales quadratically, so simply scaling the hidden size to the desired parameter count would leave the model with too few feed-forward neurons. This matters because feed-forward neurons can be interpreted as an associative memory over key--value pairs \citep{zhong2025understandingtransformerperspectiveassociative}.

\begin{figure}[t]
\centering
\includegraphics[width=\linewidth]{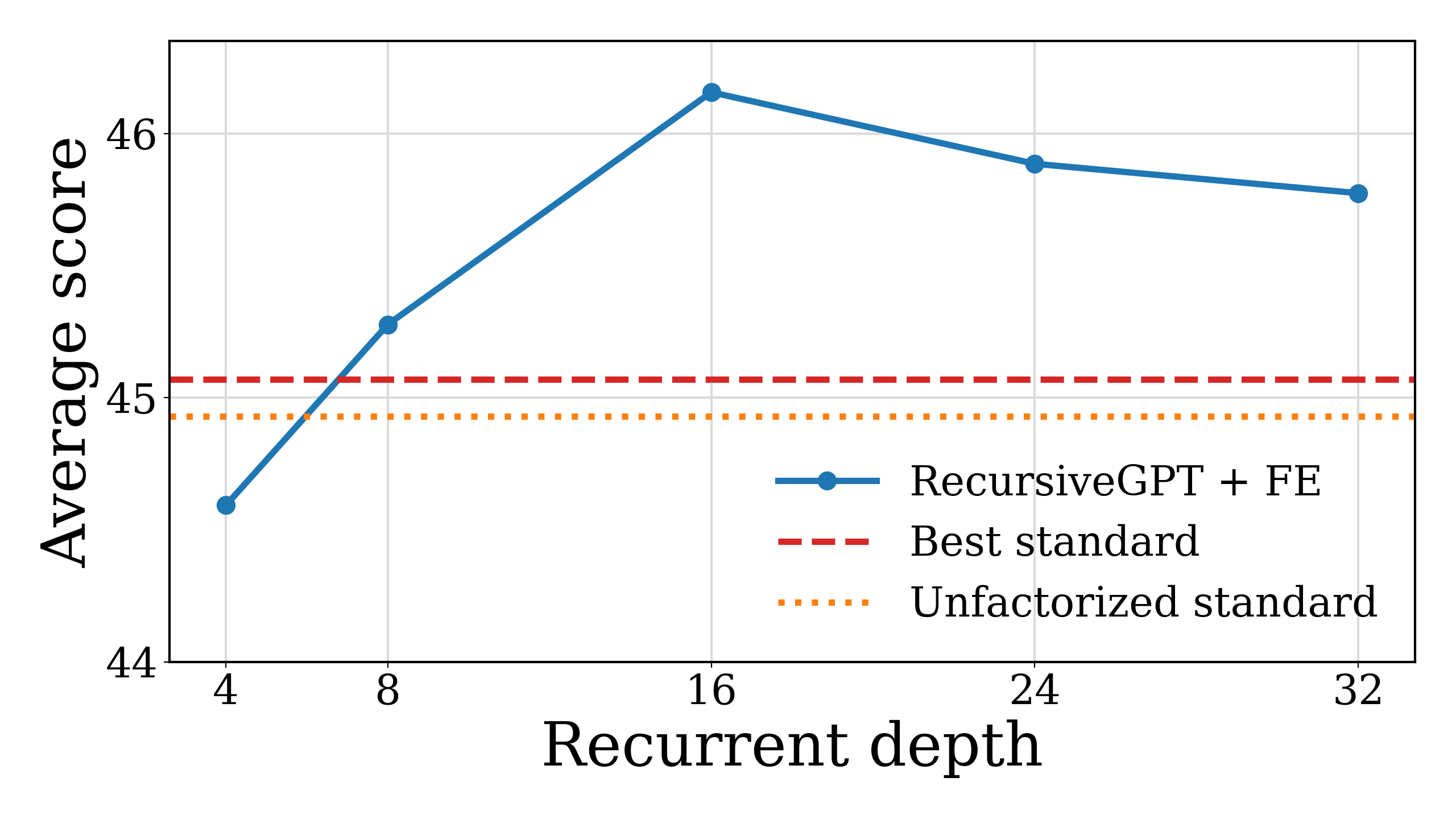}
\caption{Average accuracy for recursive models with $(H,E){=}(640,192)$ trained with recurrent depth $R$ swept from 4 to 32 on the 10\% subset of the 100M BabyLM-baseline corpus from the optimal scale experiment in Section~\ref{sec:optimal-scale}. The horizontal reference lines mark the three-seed average scores from Table~\ref{tab:std-vs-recursive} for the best standard model and an unfactorized standard model with the same hidden size.}
\label{fig:depth-sweep-lambada}
\end{figure}

\section{Results}
For the recursive experiments, we train three RecursiveGPT configurations: $(H,E){=}(640,192)$ with $R{=}16$, $(H,E){=}(1408,768)$ with $R{=}24$, and $(H,E){=}(2560,2560)$ with $R{=}24$. The first configuration is trained at 10M words, while the other two are trained at 100M words. Because of compute constraints, the 100M-word recursive models remain well below the optimal scale identified in Section~\ref{sec:optimal-scale}. These models use the same datasets as the BabyLM Challenge GPT-BERT baselines\footnote{The models compared here are not the ones introduced in the GPT-BERT paper, which do not satisfy the 10-epoch limit, but the ones submitted to BabyLM 2025.} at the corresponding budgets, so that the results are directly comparable to those baselines. Separately, for the recurrent-depth sweep and comparisons to standard models, we use the same configuration as the 10M model but train on the 10\% subset of the 100M BabyLM-baseline corpus used in the scaling experiments, enabling direct comparison to the standard-model sweep. We analyze the prediction depth of our largest configuration in Appendix~\ref{app:prediction-depth}.

\begin{table*}[t]
\centering
\scriptsize
\resizebox{\textwidth}{!}{%
\begin{tabular}{lllccccc}
\hline
Word Budget & Model & Params & BLiMP & COMPS & LAMBADA p@5 & Average \\
\hline
10M & Standard & 41.1M & 68.97 & 52.88 & 12.93 & 44.93 \\
10M & Standard + FE & 28.7M & 70.09 & 52.62 & 12.48 & 45.07 \\
10M & RecursiveGPT + FE ($R{=}16$) & 27.6M & \textbf{70.95} & \textbf{52.96} & \textbf{13.48} & \textbf{45.80} \\
\hline
100M & Standard & 1.22B & 79.71 & \textbf{60.65} & 50.04 & 63.47 \\
100M & RecursiveGPT + FE ($R{=}24$) & 124.0M & 80.06 & 58.70 & 47.01 & 61.92 \\
100M & RecursiveGPT-Large ($R{=}24$) & 404.1M & \textbf{80.70} & 60.31 & \textbf{50.77} & \textbf{63.93} \\
\hline
\end{tabular}%
}
\caption{Comparison between standard and recursive models at 10M and 100M words. FE denotes factorized embeddings and $R$ denotes recurrent depth. The 10M rows include the best standard model from Section~\ref{sec:optimal-scale} with $H{=}384$, an unfactorized $H{=}384$, $E{=}384$ model, and a depth-16 recursive model with factorized embeddings, all trained on the same 10\% subset of the 100M BabyLM-baseline corpus used in the standard-model sweep. All rows are averaged over seeds 0, 1, and 2. The 100M standard row reports the best standard model by average score. Evaluation scores are reported as percentages; Average is the arithmetic mean across the three evaluations.}
\label{tab:std-vs-recursive}
\end{table*}

\begin{table*}[t]
\centering
\small
\scriptsize
\resizebox{\textwidth}{!}{%
\begin{tabular}{lllcccccc}
\hline
Word Budget & Model & Params & BLiMP & BLiMP S. & EWoK & Entity T. & COMPS & Average \\
\hline
10M & GPT-BERT (Causal) & 31.0M & 71.70 & \textbf{63.20} & 49.50 & 34.60 & 52.80 & 54.36 \\
10M & AMLM Hard Decay & 34.7M & 71.40 & 59.20 & 51.00 & \textbf{44.20} & \textbf{54.20} & \textbf{56.00} \\
10M & RecursiveGPT + FE & 27.6M & \textbf{72.21} & 58.28 & \textbf{52.01} & 24.51 & 53.33 & 52.07 \\
\hline
100M & GPT-BERT (Causal) & 120.0M & 79.30 & 70.40 & 52.30 & 30.90 & 58.30 & 58.24 \\
100M & RecursiveGPT + FE & 124.0M & 80.06 & \textbf{70.82} & 54.74 & \textbf{34.42} & 58.70 & 59.75 \\
100M & RecursiveGPT-Large & 404.1M & \textbf{80.70} & 70.75 & \textbf{54.97} & 32.67 & \textbf{60.31} & \textbf{59.88} \\
\hline
\end{tabular}%
}

\caption{Comparison to the BabyLM 2025 GPT-BERT causal baselines \citep{charpentier-samuel-2024-bert}, the highest-scoring causal models on both tracks \citep{charpentier-etal-2025-findings}, and AMLM hard decay, the winning masked language model in the 10M-word NLP track \citep{edman-fraser-2025-mask}. RecursiveGPT scores are averaged over seeds 0, 1, and 2. Entity T. denotes Entity Tracking, and BLiMP S. denotes BLiMP Supplement. Scores are reported as percentages; Average is the arithmetic mean across the five evaluations.}
\label{tab:babylm2025-baselines}
\end{table*}

Figure~\ref{fig:depth-sweep-lambada} shows that, for the 27.6M-parameter recursive model trained on the 10\% subset of the 100M BabyLM-baseline corpus from the optimal scale experiments, the average score across the three evaluations improves with recurrent depth up to depth 16 and then declines slightly. The trend is therefore not monotonic: additional recurrent steps increase computation, but they can also make optimization more difficult. The strongest recursive setting reaches an average score of 46.16, compared with 45.07 for the best standard model from Section~\ref{sec:optimal-scale} and 44.93 for an unfactorized standard model with the same hidden size. Table~\ref{tab:std-vs-recursive} compares recursive models against standard alternatives at both 10M and 100M words. All scores are averaged over seeds 0, 1, and 2. Training-time and FLOP estimates for these models are reported in Appendix~\ref{app:compute-cost}.

Overall, the recursive models are highly competitive; in the 10M-word setting, our recursive model outperforms both standard models on all three evaluations. At 100M words, the compact 124.0M-parameter RecursiveGPT + FE model performs well on BLiMP, exceeding the 1.22B standard model while using about 10\% as many parameters, but trails it on COMPS, LAMBADA, and average score. Scaling the recursive model to 404.1M parameters yields RecursiveGPT-Large, which uses substantially higher compute and reaches the best average score among our models while still being smaller than the 1.22B standard baseline. Together, these results show that recursion provides a useful compute-scaling axis under limited data.

\begin{table*}[t]
\centering
\small
\scriptsize
\resizebox{\textwidth}{!}{%
\begin{tabular}{lccccc}
\hline
Variant & Params & BLiMP & COMPS & LAMBADA p@5 & Average \\
\hline
RecursiveGPT + FE $(H,E){=}(640,192)$ & 27.6M & 70.85 & \textbf{53.16} & \textbf{14.46} & \textbf{46.16} \\
Shared norm parameters & 27.6M & \textbf{71.38} & 52.83 & 13.97 & 46.06 \\
No Factorized Embeddings ($H{=}640$) & 56.7M & 68.13 & 52.75 & 13.81 & 44.90 \\
No Factorized Embeddings ($H{=}384$) & 30.5M & 68.70 & 52.41 & 11.17 & 44.09 \\
\hline
\end{tabular}%
}
\caption{Ablations of the 10M-word depth-16 RecursiveGPT model trained on the 10\% subset of the 100M BabyLM-baseline corpus with seed 0. FE denotes factorized embeddings. Shared norm parameters shares the normalization parameters across recurrent steps instead of using step-specific normalization and bias. No FE removes factorized input and output vocabulary maps. Scores are reported as percentages; Average is the arithmetic mean across the three evaluations.}
\label{tab:recursive-ablations}
\end{table*}

\subsection{Comparison to BabyLM 2025 winners}
\label{sec:babylm-comparison}

Table~\ref{tab:babylm2025-baselines} compares our recursive models with strong BabyLM 2025 baselines at both 10M and 100M words on BLiMP, BLiMP Supplement, EWoK, Entity Tracking, and COMPS. The baselines are the causal GPT-BERT models\footnote{Causal refers to the causal-focus model evaluated in causal mode.} and AMLM hard decay, the winning masked language model in the 10M-word NLP track. 

GPT-BERT is a hybrid causal/masked-language-modeling approach rather than a plain decoder-only baseline \citep{charpentier-samuel-2024-bert,charpentier-etal-2025-findings}. AMLM hard decay keeps a masked-language-modeling objective, but changes the masking distribution during training based on which tokens the model already predicts well \citep{edman-fraser-2025-mask}. Our goal is not to compete with these specialized recipes, but to establish RecursiveGPT as a strong architectural blueprint.

For this comparison, the recursive models use the same datasets as GPT-BERT at the corresponding budgets; the 10M model is therefore trained on the BabyLM 10M corpus rather than the 10\% subset used for the depth sweep. 

At 10M words, RecursiveGPT achieves the strongest BLiMP and EWoK scores, while AMLM achieves the strongest COMPS and average scores. The main weak point for RecursiveGPT is Entity Tracking, but we treat that benchmark with caution: even GPT-BERT scores higher on Entity Tracking at 10M words than at 100M words, which is counterintuitive for a data-scaling comparison. At 100M words, both RecursiveGPT models outperform GPT-BERT on every reported benchmark and on the overall average. RecursiveGPT-Large gives the best BLiMP, EWoK, COMPS, and average scores, while the smaller RecursiveGPT + FE model is slightly stronger on BLiMP Supplement and Entity Tracking. Overall, the recursive architecture remains competitive with strong BabyLM models for both tracks.

\subsection{Ablations}
\label{sec:recursive-ablations}

Table~\ref{tab:recursive-ablations} ablates two components of the 10M-word depth-16 RecursiveGPT model from Table~\ref{tab:std-vs-recursive}, using the same 10\% subset of the 100M BabyLM-baseline corpus. Sharing normalization parameters across recurrent steps slightly improves BLiMP, but reduces the other two evaluation scores and the overall average. Removing factorized embeddings hurts substantially, even when reducing the size of the model further to match parameter counts. The drop is larger than the corresponding gap between factorized and unfactorized standard models in Table~\ref{tab:std-vs-recursive}, suggesting that the embedding bottleneck is especially useful for the recursive model at this scale. The difference in the normalization ablation is smaller, indicating that depth conditioning is less important than we anticipated.

\subsection{Compute-Matched Standard Transformers}
\label{sec:compute-matching}
The standard-model sweep shows that, under fixed data, increasing parameter count beyond the best scale reduces downstream performance. 
Another axis of scaling compute is epoch count. We test this by matching the FLOPs of the RecursiveGPT models by training standard models for more epochs. 
We show that this results in degraded performance at both word budgets in Appendix \ref{app:additional-epochs}.

\section{Discussion}
\label{sec:discussion}
The comparisons to BabyLM winners in Section~\ref{sec:babylm-comparison} should not be read as positioning recursive Transformers against specialized training recipes or objectives. Our contribution is architectural: factorized embeddings and recursive weight sharing address parameter allocation and compute-depth coupling, providing a pathway for scaling up compute under fixed data. These architectural changes are largely orthogonal to the specialized training techniques used in GPT-BERT and AMLM hard decay. We therefore view RecursiveGPT as a useful architectural base on which more specialized data-efficient training methods can be built.

Recursive weight sharing makes optimization more delicate. Because the same block is applied many times, small changes in the recurrent transition can be amplified across depth; at the same time, if the transition is too close to an identity or a fixed point, additional recurrent steps may add little useful computation. This tension is consistent with the depth sweep in Figure~\ref{fig:depth-sweep-lambada}, where performance improves up to $R{=}16$ and then drops slightly at larger depths. It is also related to known stability issues in deep Transformers, where residual-branch dependence can amplify parameter perturbations during training \citep{liu-etal-2020-understanding}. Recent work has analyzed this stability problem by treating recursive architectures as dynamical systems over the residual stream and linking instability to residual explosion, loss spikes, and large injection-parameter spectral norms \citep{prairie2026parcae}.

Recursive Transformers also make it possible to control computation dynamically. Adaptive computation time (ACT) lets recurrent models learn when to halt \citep{graves2016adaptive}, so easier token positions could use less compute at inference. During batched Transformer training, savings are less direct because token positions are processed in parallel and the batch typically runs until the deepest active position halts. Depth scheduling is a simpler training-time alternative, starting with fewer recurrent steps and increasing depth later. As discussed in Section~\ref{sec:recursivegpt-architecture}, we do not use either technique here, but Appendix~\ref{app:prediction-depth} shows that many predictions settle before the final recurrent step, suggesting that adaptive or scheduled computation could reduce cost in future recursive models.

\section{Conclusion}
We studied language model pre-training in a data-limited, compute-abundant regime. Across 10M--100M word budgets, two corpora, and multiple evaluations, standard decoder-only Transformers often show non-monotonic performance with parameter count, and the best model size depends on both the data budget and the downstream target.

This behavior reflects an architectural mismatch. Under a fixed data budget, the usual ways of giving a standard Transformer more compute also add trainable capacity, and pushing along this coupled scaling axis eventually hurts performance. RecursiveGPT addresses this mismatch by turning recurrent depth into a compute-scaling axis that is decoupled from parameter count. This lets us scale up compute and extract more performance from the available data at both budgets we evaluate.

RecursiveGPT is also competitive with strong BabyLM 2025 models, improving over the causal GPT-BERT baseline on all 100M-word evaluations and remaining close to the best 10M-word models despite using a simple causal objective. More broadly, our results suggest that limited-data pre-training should be treated as its own scaling regime: rather than simply shrinking web-scale architectures, we should develop methods whose inductive biases and scaling behavior are better matched to the constraints of limited data.

\subsection{Future Work}
An immediate next step is to combine recursive architectures with the specialized objectives and training recipes that have proven effective in BabyLM. The design space of recursion itself is also much broader than the single-block, fixed-depth models studied here. Adaptive halting, depth scheduling, partial sharing, and richer depth-conditioning mechanisms could preserve the benefits of recurrence while reducing training and inference cost.

\section*{Limitations}
Our hyperparameter tuning for experiments was limited. We tuned learning rates, optimizer settings, and architectural details locally to obtain stable and competitive runs, but did not retune hyperparameters separately for every combination of corpus, data budget, model size, and recurrent depth. Such a full search would be computationally infeasible with our resources.

Our study also covers only a small part of the recursive Transformer design space. Our models use a single shared block, a fixed number of recurrent steps, and simple per-depth conditioning parameters. Without efficiency mechanisms such as adaptive computation time or depth scheduling, increasing recurrent depth substantially increases training and inference cost.

\bibliography{references}

\clearpage
\appendix

\twocolumn[{
\begin{center}
\includegraphics[width=\textwidth]{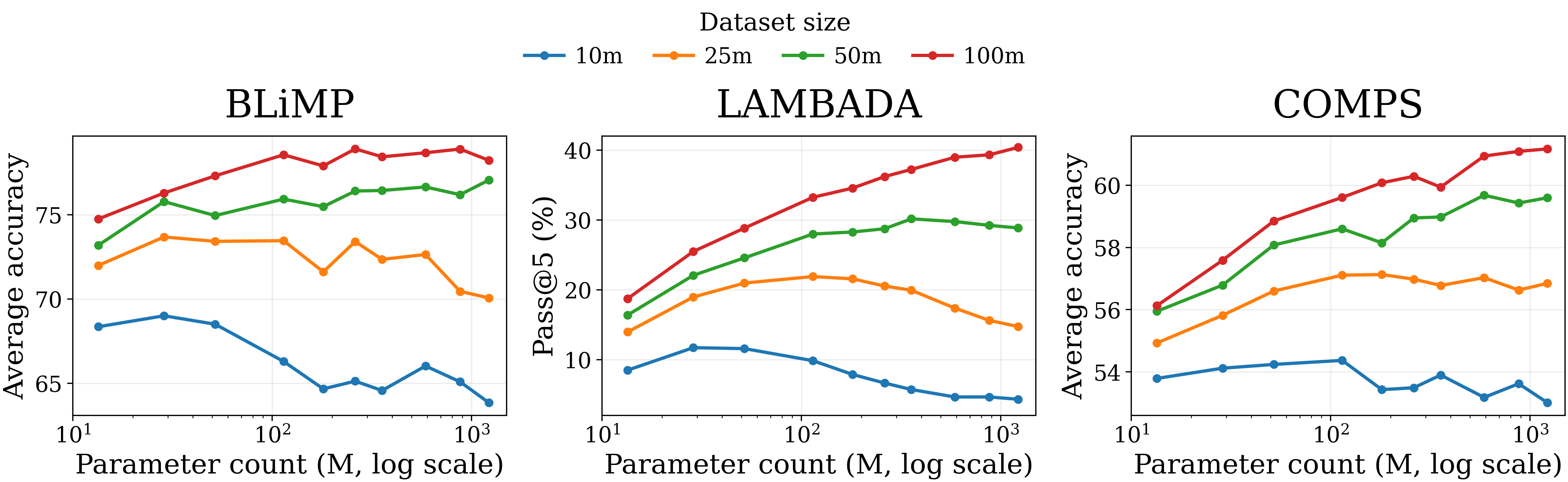}
{\captionsetup{hypcap=false}\captionof{figure}{Optimal scale on ClimbMix across the same three evaluations shown in Figure~\ref{fig:optimal-scale}. As in the BabyLM baseline corpus, the preferred parameter count depends on both dataset size and evaluation target.}\label{fig:optimal-scale-climbmix}}
\end{center}
}]

\section{ClimbMix Scaling Curves}
Figure~\ref{fig:optimal-scale-climbmix} shows the same three evaluations on ClimbMix. Preferred parameter ranges remain broadly stable, while absolute scores shift: the best BLiMP scores stay within 1.2 points of those on the BabyLM-baseline corpus, the best COMPS scores are consistently higher, with a maximum gain of 1.6 points, and LAMBADA pass@5 is lower by 1.9 points at 10M words and 9.4 points at 100M words. Thus, the dependence on data budget and evaluation target remains qualitatively similar across the two corpora.

\section{Architecture Details}
\label{app:architecture-details}
The standard parameter/data sweep uses a 12-layer decoder-only causal Transformer architecture following \citet{10.5555/3295222.3295349}. Each model consists of independent Transformer blocks, with no recursive weight sharing or mixture-of-experts layers. Each block uses a pre-norm residual structure, causal self-attention, and a dense two-layer $ReLU^2$ feed-forward sublayer.

Attention uses fused QKV projections with an attention head size of 64, rotary position embeddings (RoPE; \citet{10.1016/j.neucom.2023.127063}), and query-key normalization \citep{henry-etal-2020-query}. To support packed training without allowing tokens to attend across document boundaries, we use variable-length causal attention computed with FlashAttention 2 kernels \citep{dao2024flashattention}. We also apply a learned per-head sigmoid gate to the scaled dot-product attention output, following gated attention for LLMs \citep{qiu2026gated}. This is intended to reduce reliance on attention sinks, since our setup does not insert a BOS token or other special token that the model could use as an attention sink.

Output projections in both the attention and MLP sublayers are zero-initialized. Embeddings are untied from the language-model head; where used, both input embeddings and output heads follow the ALBERT-style factorized projection idea \citep{DBLP:conf/iclr/LanCGGSS20}. We use a vocabulary size of 32768 with a BPE tokenizer trained using rustbpe \citep{karpathy2025rustbpe} on the corresponding full corpus used to construct the training subset.

Training used a two-group optimizer. Transformer block parameters were optimized with Muon \citep{jordan2024muon} augmented with neuron-wise adaptive learning rates \citep{li2025normuonmakingmuonefficient}, using learning rate 0.02, momentum 0.95, second-moment coefficient 0.95, and weight decay 0.1. Embeddings, normalization parameters, and other auxiliary parameters were optimized with Adam \citep{Kingma2014AdamAM} using learning rate 0.005, $\beta=(0.9,0.95)$, $\epsilon=10^{-10}$, and weight decay 0.005. In both groups we used cautious weight decay \citep{chen2026cautious}, applying decay only to coordinates where the parameter and update directions are aligned. Learning rates used 50 warmup steps, followed by a constant phase and then a linear cooldown to 0 over the final 20\% of training steps. Gradients were clipped to norm 2.0 and training used bfloat16 autocasting. All runs used an effective global batch size of 32768 tokens and random seed 0. Multi-seed runs used seeds 0, 1, and 2.
All individual models were trained on a single NVIDIA A100 80GB GPU, except for the 100M-word RecursiveGPT models, which were trained on four such GPUs.

\twocolumn[{
\begin{center}
\includegraphics[width=.92\textwidth]{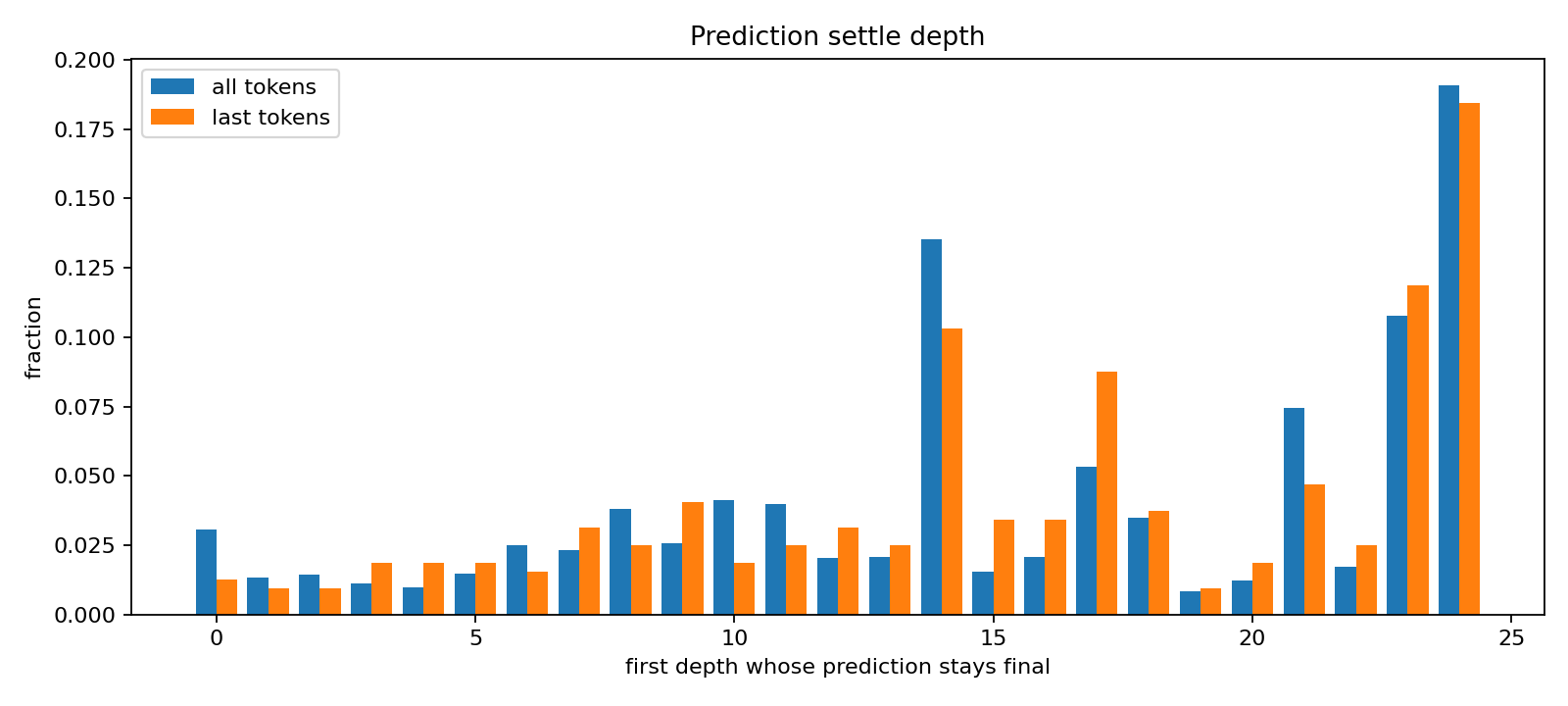}
{\captionsetup{hypcap=false}\captionof{figure}{Histogram of settle depths for the depth-24 RecursiveGPT-Large model. ``All tokens'' includes all 65{,}536 analyzed token positions; ``last tokens'' includes only the final token of each sampled document.}\label{fig:settle-depth-histogram}}
\vspace{-.25em}
\includegraphics[width=.92\textwidth]{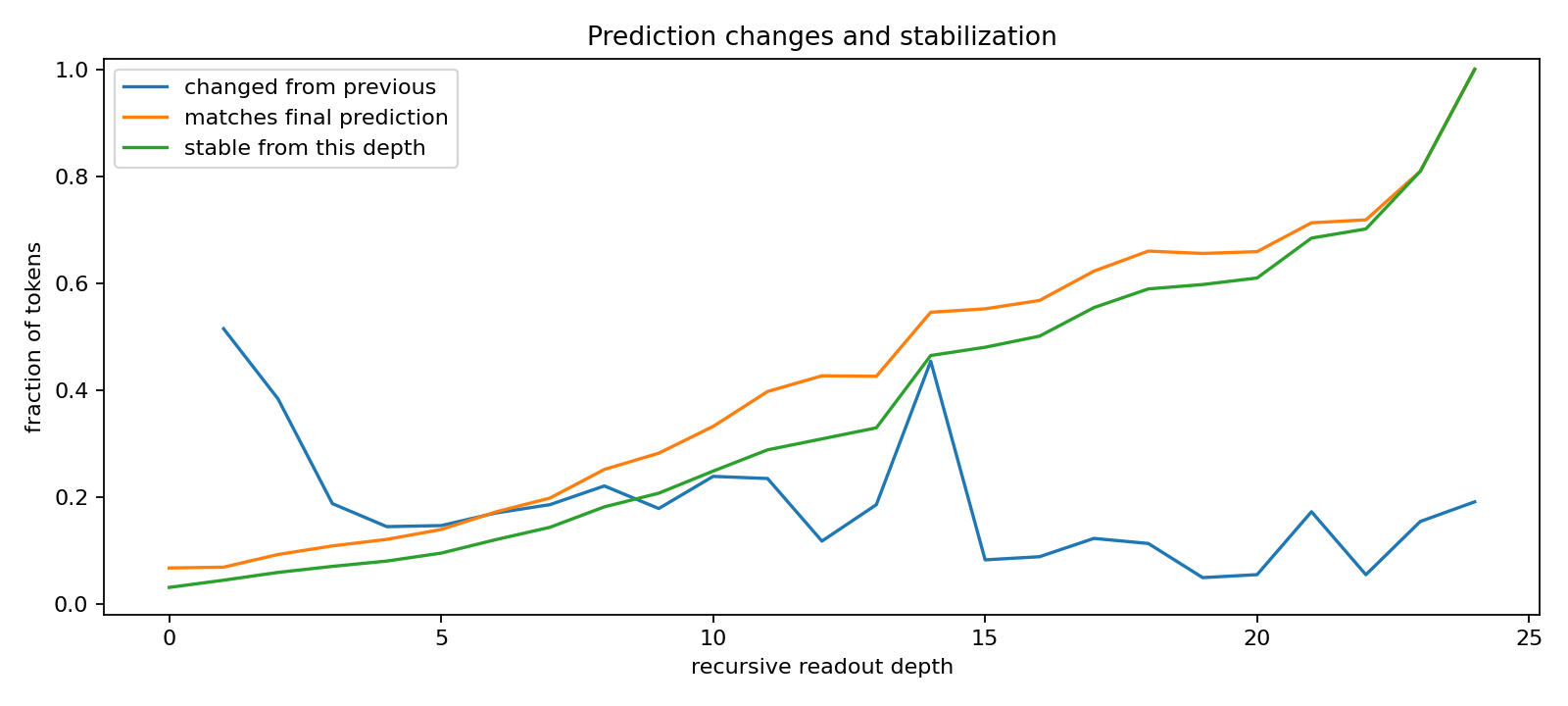}
{\captionsetup{hypcap=false}\captionof{figure}{Prediction dynamics across recurrent depth. Final agreement measures whether the current top-1 prediction matches the depth-24 prediction; stable from depth measures whether it matches the depth-24 prediction and stays fixed through all later depths.}\label{fig:prediction-dynamics}}
\end{center}
}]

\section{Prediction Depth and Adaptive Computation}
\label{app:prediction-depth}
We analyze the $R{=}24$ RecursiveGPT-Large model trained on 100M words using 65{,}536 tokens from randomly chosen training documents. At each recurrent depth, we read the intermediate hidden states through the model's own output head and record the resulting top-1 predictions (analogous to the logit lens; \citealp{nostalgebraist2020logitlens}). For each token position, we define its settle depth as the earliest recurrent depth whose top-1 prediction matches the final depth-24 prediction and remains unchanged thereafter. We report results for all token positions, and separately for the last token in each document.

Figure~\ref{fig:settle-depth-histogram} shows that, even without ACT or an explicit incentive to predict early, the model naturally learns to lock in its final prediction at particular recurrent depths. The largest masses for all tokens are at depths 24 (19.1\%), 14 (13.5\%), and 23 (10.8\%), with a similar pattern for last-token positions.

Figure~\ref{fig:prediction-dynamics} suggests that adaptive halting could save substantial inference compute. Among all tokens, about half are already stable by depth 16, 70.1\% by depth 22, and 80.9\% by depth 23. Thus roughly four out of five tokens settle before the final recurrent step, and only 29.9\% need the last two depths.

\twocolumn[{
\begin{center}
\scriptsize
\resizebox{\textwidth}{!}{%
\begin{tabular}{lllccc}
\hline
Word Budget & Model & Params & $N_{\mathrm{eff}}$ & A100-time & Est. train FLOPs \\
\hline
10M & Standard & 41.1M & 15.97M & 9 Minutes & 1.25e16 \\
10M & Standard + FE & 28.7M & 16.12M & 9 Minutes & 1.26e16 \\
10M & RecursiveGPT + FE ($R{=}16$) & 27.6M & 236.32M & 21 Minutes & 1.84e17 \\
\hline
100M & Standard & 1.22B & 1.021B & 17 hours 33 Minutes & 7.96e18 \\
100M & RecursiveGPT + FE ($R{=}24$) & 124.0M & 1.716B & 22 Hours 36 Minutes & 1.34e19 \\
100M & RecursiveGPT-Large ($R{=}24$) & 404.1M & 5.665B & 64 Hours 24 Minutes & 4.42e19 \\
\hline
\end{tabular}%
}
{\captionsetup{hypcap=false}\captionof{table}{Compute costs for the models in Table~\ref{tab:std-vs-recursive}. Params is the total number of trainable parameters, including the input and output vocabulary maps. $N_{\mathrm{eff}}$ is the non-vocabulary parameter count effectively executed per token. Estimated training FLOPs use $6N_{\mathrm{eff}}T$ \citep{kaplan2020scalinglawsneurallanguage}. A100 time excludes initial compilation and is summed over devices for multi-GPU runs.}\label{tab:compute-cost}}\par\smallskip
\end{center}
}]

\section{Compute Cost}
\label{app:compute-cost}
For fixed model dimensions, compute scales approximately linearly with recurrent depth:
\[
F(R) \approx F_{\mathrm{fixed}} + R F_{\mathrm{block}}.
\]
We make this relationship explicit below through the effective parameter count and the $6NT$ FLOP estimate.

Table~\ref{tab:compute-cost} distinguishes the total parameter count, which includes the input and output vocabulary maps, from $N_{\mathrm{eff}}$, the parameter count used in the FLOP estimate. Following the non-vocabulary convention of \citet{kaplan2020scalinglawsneurallanguage}, $N_{\mathrm{eff}}$ excludes the vocabulary maps but includes the dense embedding-to-hidden and hidden-to-embedding projections.

We estimate training FLOPs as $6N_{\mathrm{eff}}T$. All models are trained for 10 epochs, giving $T{\approx}130$M BPE tokens for the 10M-word setting and $T{\approx}1.3$B for the 100M-word setting. For standard models, $N_{\mathrm{eff}}$ counts every non-vocabulary parameter once. For recursive models, we use
\[
N_{\mathrm{eff}}
= R N_{\mathrm{shared\ block}}
+ N_{\mathrm{step\text{-}specific}}
+ N_{\mathrm{fixed}},
\]
so only the shared block is multiplied by $R$; step-specific normalization and bias parameters and fixed projections are counted once. The $6NT$ values are approximate, so Table~\ref{tab:compute-cost} also reports measured A100 time.

\par\newpage
\section{Factorized-Embedding Sweep}
\label{app:factorized-embedding-sweep}
\begin{table}[!h]
\centering
\scriptsize
\setlength{\tabcolsep}{3pt}
\resizebox{\linewidth}{!}{%
\begin{tabular}{rcccccc}
\hline
$E$ & Embed & Total & BLiMP & COMPS & LAMBADA & Average \\
\hline
64 & 4.2M & 20.2M & \textbf{70.51} & 51.96 & 9.79 & 44.09 \\
128 & 8.5M & 24.5M & 69.79 & 52.33 & 12.49 & 44.87 \\
192 & 12.7M & 28.7M & 70.09 & 52.62 & 12.48 & \textbf{45.07} \\
256 & 17.0M & 32.9M & 69.02 & 52.55 & 12.43 & 44.67 \\
320 & 21.2M & 37.2M & 69.25 & 52.31 & \textbf{13.01} & 44.86 \\
384 & 25.2M & 41.1M & 68.97 & \textbf{52.88} & 12.93 & 44.93 \\
\hline
\end{tabular}%
}
\caption{Factorized-embedding sweep on the 10M-word setting with hidden size fixed at $H{=}384$. Embed and Total report vocabulary-map (input embedding plus output head) and total parameter counts; scores are percentages averaged over seeds 0, 1, and 2, LAMBADA is evaluated as pass@5, and Average is the arithmetic mean of the scores.}
\label{tab:fe-sweep}
\end{table}
In Table~\ref{tab:fe-sweep}, we evaluate factorized embeddings by sweeping $E$ while keeping $H = 384$ fixed for a standard model trained on 10M words, reporting averages over seeds 0, 1, and 2. The final row at $E = 384$ is the unfactorized input/output vocabulary-map case. This sweep does not identify a precise optimum, and the differences between nearby settings are small. $E{=}192$ gives the highest average score, $E{=}64$ gives the strongest BLiMP score, $E{=}320$ gives the strongest LAMBADA pass@5 score, and unfactorized gives the strongest COMPS score. We therefore identify $E{=}192$ as a balanced choice rather than a definitive optimum. In the RecursiveGPT ablations in Section~\ref{sec:recursive-ablations}, we find that the benefit from factorized embeddings is more substantial, suggesting that this effect becomes clearer as we scale up compute.

\clearpage
\twocolumn[{
\begin{center}
\scriptsize
\resizebox{\textwidth}{!}{%
\begin{tabular}{lllrrcccc}
\hline
Budget & Model & Params & Epochs & Est. train FLOPs & BLiMP & COMPS & LAMBADA & Average \\
\hline
10M & Standard + FE & 28.7M & 10 & 1.26e16 & 70.09 & 52.62 & 12.48 & 45.07 \\
10M & Standard + FE & 28.7M & 146 & 1.84e17 & 63.11 & 51.31 & 10.00 & 41.47 \\
10M & RecursiveGPT + FE ($R{=}16$) & 27.6M & 10 & 1.84e17 & \textbf{70.95} & \textbf{52.96} & \textbf{13.48} & \textbf{45.80} \\
\hline
100M & Standard & 1.22B & 10 & 7.96e18 & 79.71 & \textbf{60.65} & 50.04 & 63.47 \\
100M & Standard & 1.22B & 50 & 3.98e19 & 74.63 & 58.74 & 36.68 & 56.68 \\
100M & RecursiveGPT-Large ($R{=}24$) & 404.1M & 10 & 4.42e19 & \textbf{80.70} & 60.31 & \textbf{50.77} & \textbf{63.93} \\
\hline
\end{tabular}
}
{\captionsetup{hypcap=false}\captionof{table}{Standard Transformers trained for additional epochs compared with the original 10-epoch baselines and recursive models. Estimated training FLOPs follow the calculation in Appendix~\ref{app:compute-cost}. Scores are percentages averaged over seeds 0, 1, and 2; LAMBADA is pass@5, and Average is the arithmetic mean of BLiMP, COMPS, and LAMBADA.}\label{tab:additional-epochs}}\par\smallskip
\end{center}
}]

\section{Compute Matching with Additional Epochs}
\label{app:additional-epochs}
To test whether the additional compute used by RecursiveGPT can instead be allocated to repeated passes over the training corpus, we train the selected standard baselines for additional epochs. Table~\ref{tab:additional-epochs} compares them with the original 10-epoch baselines and recursive models on the three evaluations used in our main comparison. At 10M words, we train for 146 epochs, approximately matching the estimated training FLOPs of RecursiveGPT. At 100M words, we train for 50 epochs.

Additional epochs reduce the average score from 45.07 to 41.47 at 10M words and from 63.47 to 56.68 at 100M words. This degradation is consistent with overfitting and shows that, in our experiments, neither increasing standard-model size nor allocating additional compute through repeated passes over the fixed data matches the performance of RecursiveGPT.

\end{document}